\pdfoutput=1
\documentclass[11pt]{article}

\usepackage[final]{acl}

\usepackage{times}
\usepackage{latexsym}
\usepackage[T1]{fontenc}
\usepackage[utf8]{inputenc}
\usepackage{microtype}
\usepackage{inconsolata}
\usepackage{graphicx}
\usepackage{booktabs}
\usepackage{multirow}

\usepackage{xcolor}
\usepackage[table]{xcolor}
\usepackage{tabularx}
\usepackage{listings}
\usepackage{booktabs}
\usepackage{tcolorbox}

\usepackage{pifont}
\newcommand{\cmark}{\ding{51}} % ✔
\newcommand{\xmark}{\ding{55}} % ✗
\usepackage{CJKutf8}
\usepackage{algorithm}
\usepackage{algorithmic}
\usepackage{fontawesome}
\usepackage{fancyhdr}
\fancypagestyle{firstpage}{
  \fancyhf{}
  \lhead{\raisebox{-0.2\height}[0pt][0pt]{%
    \makebox[0pt][l]{%
      \includegraphics[width=0.3\textwidth]{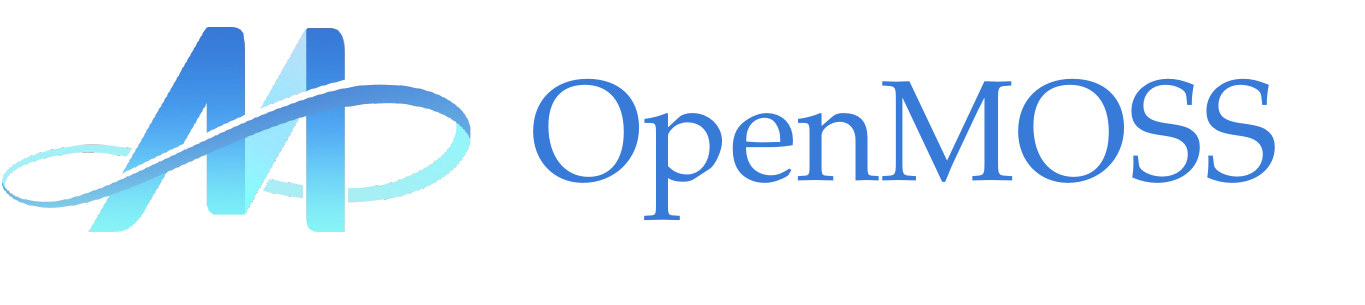}%
    }%
  }}
  
  \setlength{\headwidth}{\textwidth}
}

\title{WESR: Scaling and Evaluating Word-level Event-Speech Recognition}

\author{
Chenchen Yang$^{1,2}$\hspace{.3em}
Kexin Huang$^{1}$ \hspace{.3em}
Liwei Fan$^{1}$ \hspace{.3em}
Qian Tu$^{1}$ \hspace{.3em}
\\
\textbf{
Botian Jiang$^{1,2}$ \hspace{.3em}
Dong Zhang$^{1}$ \hspace{.3em}
Linqi Yin$^{1,2} $ \hspace{.3em}
Shimin Li$^{1}$ \hspace{.3em}
}
\\
\textbf{
Zhaoye Fei$^{1}$ \hspace{.3em}
Qinyuan Cheng$^{1}$ \hspace{.3em}
Xipeng Qiu$^{1,2}$\thanks{ {} Corresponding author.}
}
\\
\texttt{ccyang6971@gmail.com, chengqy21@m.fudan.edu.cn} \\ 
[1ex]
$^{1}$Fudan University \\
$^{2}$Shanghai Innovation Institute \\
}

\begin{document}

\thispagestyle{firstpage}

\maketitle

\pagestyle{plain}

\begin{abstract}

Speech conveys not only linguistic information but also rich non-verbal vocal events such as laughing and crying. While semantic transcription is well-studied, the precise localization of non-verbal events remains a critical yet under-explored challenge. 
Current methods suffer from insufficient task definitions with limited category coverage and ambiguous temporal granularity. They also lack standardized evaluation frameworks, hindering the development of downstream applications.
To bridge this gap, we first develop a refined taxonomy of 21 vocal events, with a new categorization into discrete (standalone) versus continuous (mixed with speech) types. Based on the refined taxonomy, we introduce \textbf{WESR-Bench}, an expert-annotated evaluation set (900+ utterances) with a novel position-aware protocol that disentangles ASR errors from event detection, enabling precise localization measurement for both discrete and continuous events. We also build a strong baseline by constructing a 1,700+ hour corpus, and train specialized models, surpassing both open-source audio-language models and commercial APIs while preserving ASR quality. We anticipate that WESR will serve as a foundational resource for future research in modeling rich, real-world auditory scenes.
\footnote{ {} Available at \href{https://github.com/Cr-Fish/WESR}{\faGithub\ WESR}. }

\end{abstract}

\begin{figure}[t]
\centering
\includegraphics[width=1\linewidth]{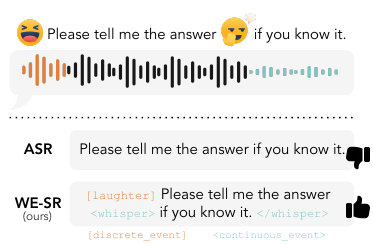}
\caption{Task overview of our WESR that generates transcripts with explicit word-aligned continuous/discrete event tags.}
\label{fig:top}
\end{figure} 

\begin{table*}[t]
    \centering
    \footnotesize
    \setlength{\tabcolsep}{4.2pt} 
    \renewcommand{\arraystretch}{1.2}
    \begin{tabularx}{\textwidth}{l l r r c c l r}
        \toprule
        \textbf{Dataset} & \textbf{Lang.} & \textbf{\# Cat.} & \textbf{Dur.~(h)} & \textbf{Cont.} & \textbf{Multi.} & \textbf{Source} & \textbf{Annotation} \\
        \midrule
        NonverbalTTS~\citep{borisov2025nonverbaltts} & EN & 10 & 17 & \xmark & \cmark & VoxCeleb, Expresso & Pipeline \\
        % \midrule
        NVSpeech-170k~\citep{liao2025nvspeech170k} & EN, ZH & 18 & 732 & \xmark & \cmark & miHoYo, Emilia & Model \\
        % \midrule
        NonVerbalSpeech-38K~\citep{ye2025nonverbalspeech38k} & EN, ZH & 10 & 131 & \cmark & \xmark & Web & Pipeline \\
        % \midrule
        SMIIP-NV~\citep{wu2025smiip} & ZH & 3 & 33 & \xmark & \cmark & In-house Recorded & -- \\
        % \midrule
        Synparaspeech~\citep{bai2025synparaspeech} & ZH & 6 & 119 & \xmark & \xmark & Synthesized & -- \\
        % \midrule
        MNV-17~\citep{mai2025mnv17} & ZH & 17 & 8 & \xmark & \xmark & In-house Recorded & -- \\
        % \midrule
        CapSpeech-SEDB~\citep{wang2025capspeech} & EN & 10 & $<\!1$ & \cmark & \xmark & Synthesized & -- \\
        \midrule
        \rowcolor{gray!12}
        \textbf{WESR-Bench} & EN, ZH & \textbf{21} & \textbf{3} & \cmark & \cmark & Web & Human \\
        \rowcolor{gray!12}
        \textbf{WESR-Train} & EN, ZH & \textbf{21} & \textbf{1,767} & \cmark & \cmark & Web & Gemini \\
        \bottomrule
    \end{tabularx}
    \caption{
        Comparison of existing non-verbal speech datasets. ``Lang.'' denotes language coverage. ``\# Cat.'' indicates the number of included categories. ``Dur.'' denotes the total duration. ``Cont.'' indicates whether the dataset contains continuous events. ``Multi.'' indicates whether an utterance can be annotated with multiple event categories. ``--'' indicates that annotations are inherently available from the recording or synthesis process.
    }
    \label{tab:dataset_comparison}
\end{table*}

\section{Introduction}
Speech, a crucial component of human communication, not only conveys textual information but also embeds rich non-verbal cues, such as speaker emotions, intonation, and diverse speech events like laughter, coughing, and whispering. These events reflect the speaker’s intent and expressive style, significantly enhancing the contextual meaning and expressiveness of speech. 

However, simply knowing that an event occurred is not enough; knowing exactly \textit{where} it happened is just as important for understanding the meaning. Sentence-level detection often misses the specific relationship between the words and the non-verbal sounds. For instance, consider the difference between \textit{``<laughing> I can't believe you did that! </laughing>''}, which sounds like a friendly joke, versus \textit{``I can't believe you did <laughing> that </laughing>''}, where the laughter on a specific word might imply mockery or disbelief about the action itself. Even though the lexical content is the same, the position of laughing changes the meaning completely. Therefore, we need fine-grained, word-level modeling to capture these subtle differences and truly understand what the speaker means.

Despite this, achieving word-level vocal event transcriptions remains a significant challenge. Conventional automatic speech recognition (ASR) and vocal event detection (VED) systems cannot robustly perform such a task. On one hand, most ASR models focus solely on converting speech to plain text, ignoring non-verbal events as noise and thus losing crucial context. On the other hand, conventional VED datasets and methods operate at the utterance or frame level classification, lacking fine-grained alignment with spoken words. 

Recent work has introduced several word-level vocal event datasets for event-aware ASR, but they either cover only a small set of event categories, contain limited hours of audio, or exhibit highly imbalanced language proportions. In addition, while some corpora include continuous event tags, they typically treat them as overlapping sound events rather than as vocalizations that modulate the speech itself. 

On the evaluation side, current event-aware ASR systems are usually assessed with word error rate (WER) together with sentence-level event classification accuracy, without explicitly accounting for the word-level positions of the events. A few studies further incorporate penalties for positional misalignment of tags within the sentence, but their metrics still cannot naturally handle multiple events occurring in the same utterance. There are also no effective evaluation methods for continuous events that span multiple words.

To address these gaps, we propose \textbf{Word-level Event-Speech Recognition~(WESR)}. On the evaluation side, we introduce WESR-Bench, an expert-annotated evaluation set of 900+ utterances spanning 21 event categories (15 discrete, 6 continuous), along with a robust, position-aware evaluation protocol that decouples lexical errors from event localization and jointly scores event type and word-level alignment, with native support for multiple events in one sentence. On the modeling side, we train WESR models by constructing WESR-Train, a large-scale corpus totaling 1,700+ hours of speech with word-level event transcriptions. Through extensive experiments, we demonstrate the efficacy of our data-centric approach: our resulting models outperform strong open-source audio understanding baselines as well as commercial APIs, while preserving ASR quality. Together, our model and evaluation framework establish a reliable foundation for training and benchmarking event-aware ASR in bilingual, naturalistic speech.

% 4. ours
In summary, our main contributions are:

\begin{itemize}
    \item \textbf{Comprehensive Definition}: We formalize the Word-level Event–Speech Recognition (WESR) task by establishing a rigorous taxonomy of 21 vocal event categories and distinguishing between \textit{discrete} and \textit{continuous} attributes, providing a comprehensive framework for modeling paralinguistic information alongside lexical content.
    
    \item \textbf{Benchmarking}: We introduce \textsc{WESR-Bench}, a 900+ samples, expert-annotated test set featuring bilingual, naturalistic speech based on our taxonomy. 
    
    \item \textbf{A Strong Baseline for WESR}: We construct a WESR-specialized baseline that surpasses strong open-source audio language models and proprietary APIs on the WESR, serving as a convenient tool for the community.
    
\end{itemize}

\section{Related Works}
\subsection{Automatic Speech Recognition}
ASR technology has achieved remarkable progress in recent years, driven by large-scale pre-training and multi-task learning. Notable models such as OpenAI’s Whisper~\cite{radford2023whisper} support multilingual recognition and a variety of tasks including speech translation and alignment, demonstrating strong cross-lingual generalization. Similarly, models like SeedASR~\cite{bai2024seedasr} and FireRedASR~\cite{xu2025fireredasr} are widely adopted in both industry and academia. Despite their impressive performance in linguistic content recognition, these models generally overlook non-verbal events and environmental information in speech. 

\subsection{Vocal Event Detection}
Vocal event detection was initially a classification task, aiming to determine whether an audio segment contains specific types of vocal event. Early datasets such as ESC-50~\cite{piczak2015esc} and VocalSound~\cite{gong2022vocalsound} provide clear category labels, enabling single- or multi-label model training. These tasks typically focus on utterance-level event discrimination and struggle to accurately identify the temporal location of events. AudioSet~\cite{gemmeke2017audioset} expanded the task by introducing large-scale, multi-label weakly tagged audio and, in 2021, provided strong labels with precise timestamps, supporting temporal-aware event detection. Most existing methods, such as HTS-AT~\cite{chen2022hts}, define event detection as utterance-level classification, while others like PANNs~\cite{kong2020panns} support frame-level outputs with temporal information. However, these approaches lack integration with semantic information and joint modeling of the relationship between speech content and events.

\subsection{Non-Verbal Speech Corpora}
A series of recent datasets focuses on the inline modeling of non-verbal events. Some methods produce word-level labels through a pipeline (frame-level event detection followed by word alignment to insert events into the transcription), such as NonverbalTTS~\citep{borisov2025nonverbaltts} and NonVerbalSpeech-38K~\citep{ye2025nonverbalspeech38k}. This approach ensures that event-containing speech segments can be mined from large-scale corpora, but the labeling accuracy and label coverage is limited by the event detection model, and the multi-stage process (event classification, ASR, and alignment) leads to error accumulation, while inherently limited by the performance of the annotation model~\citep{mai2025mnv17}.

Other methods, such as SynParaSpeech~\citep{bai2025synparaspeech} and CapSpeech~\citep{wang2025capspeech}, obtain event-annotated data through synthesis. SynParaSpeech uses voice conversion to transform vocal event clips, while the speech portion is generated by TTS; CapSpeech relies on human annotators to insert event audio into speech recordings. These methods can achieve precise word-level labels (the insertion positions are explicitly specified), but they tend to suffer from unnaturalness in the resulting audio.

There are also approaches that rely on manual annotation. NVSpeech-170k~\citep{liao2025nvspeech170k} manually labeled 48k audio clips covering 18 event categories, then trained an ASR model on this annotated subset and used the model to automatically expand the dataset to 170k samples. SMIIP-NV~\citep{wu2025smiip} and MNV-17~\citep{mai2025mnv17} recruited participants to record speech with non-verbal events. SMIIP-NV collected 33 hours of manually recorded data with 3 event categories, and MNV-17 collected 7.55 hours of manually recorded Chinese data with 17 categories.

However, these methods share several common limitations: 1) the event category design is either too small or contains many only marginally distinctive tags (a summary of tags used in prior work is provided in Supplementary Table~\ref{tab:prior_tags}); 2) the amount of high-quality annotated data for robust evaluation remains limited; 3) continuous events are poorly defined—even when datasets include tags such as \texttt{<B>...</B>} to indicate continuity, they typically denote overlapping sounds rather than nonverbal speech coupled with the transcript; and 4) there is no widely accepted evaluation protocol: most event-aware ASR systems are evaluated only with classification F1 and WER/CER, which ignore word-level localization, while metrics such as TPD/NTD in NonVerbalSpeech-38K cannot handle multiple or continuous events within the same sentence, and utterance-level accuracy in MNV-17 obscures per-category behavior and still fails to capture word-level performance. Comparison of all datasets is shown in Table~\ref{tab:dataset_comparison}.

\begin{table*}[t]
\centering
\fontsize{8}{12}\selectfont
\begin{tabularx}{\textwidth}{
    >{\raggedright\arraybackslash}X
    >{\raggedright\arraybackslash}p{3cm}
    >{\raggedright\arraybackslash}l
}
\toprule
\textbf{Example in WESR-Bench} & \textbf{Tags} & \textbf{Type} \\
\midrule
\begin{CJK}{UTF8}{gbsn}
好的，\texttt{[clear\_throat]}那我们先来说一个比较轻松的话题，嗯，可能。
\end{CJK} 
& \texttt{[clear\_throat]} & Discrete \\
\midrule
Alexander! \texttt{[laughs]} Oh, my little warrior. Come here. Come on. \texttt{[laughs]}
& \texttt{[laughs]} & Discrete \\
\midrule
\begin{CJK}{UTF8}{gbsn}
\texttt{<laughing>}诶，不是重点，然后我就想说\texttt{</laughing>}，那我就可以稍微减少一点点儿，就是碳水，然后呢可能多吃一点儿这个蛋白，然后多吃纤维。
\end{CJK}
& \texttt{<laughing>} & Continuous \\
\midrule
\texttt{<singing>} I wish you a Merry Christmas, I wish you a Merry Christmas and a Happy New Year. \texttt{</singing>}
& \texttt{<singing>} & Continuous \\
\midrule
\begin{CJK}{UTF8}{gbsn}
\texttt{<shouting>}住手,快点住手\texttt{</shouting>}\texttt{[giggle]}那我走了。
\end{CJK}
& \texttt{<shouting>}, \texttt{[giggle]} & Mixed \\
\midrule
\texttt{<crying>} Oh my face, my face. \texttt{</crying>} \texttt{<shouting>} I brought sin into this world once. \texttt{[inhale]} I couldn't risk it again. \texttt{</shouting>} \texttt{[sobbing]}
& \texttt{<crying>}, \texttt{<shouting>}, \texttt{[inhale]}, \texttt{[sobbing]} & Mixed \\
\bottomrule
\end{tabularx}
\caption{Examples from WESR-Bench demonstrating three event types: discrete events, continuous events, and mixed scenarios containing multiple event types within a single utterance.}
\label{tab:casestudy}
\end{table*}

\section{Task Formulation}
To facilitate standardized evaluation of paralinguistic modeling, we define the task of WESR. Given an input audio stream, the objective of the WESR task is to generate a transcription that not only contains the spoken content (as in conventional ASR), but also annotates non-verbal vocal event tags at the correct word-level position of the transcription. Specifically, the output transcription should include: 1) \textbf{Discrete event tags:} These tags mark brief, discrete events such as \texttt{[laughs]} and \texttt{[clear\_throat]}, which occur at specific time points within the audio. 2) \textbf{Continuous event tags:} These capture vocal events that modulate the speech itself, such as speaking while laughing, or singing. Represented by pairs like \texttt{<laughing>... </laughing>} and \texttt{<singing>... </singing>}.

For category selection, we start from a broad list of vocal events reported in prior work and commonly observed in conversational media (e.g., podcasts, livestreams, audiobooks). We further refine this set through pilot annotation on a held-out subset of data, discarding labels that annotators find ambiguous or inconsistent, and arrive at a compact yet expressive taxonomy that covers both discrete events and continuous events while remaining feasible for large-scale labeling and reliable evaluation. The entire tag taxonomy is detailed in Supplementary Table~\ref{tab:label_aggregation}.

Formally, given an input audio segment $x$, the model outputs a sequence $Y = (y_1, y_2, \ldots, y_n)$, where each $y_i$ is either a word from the spoken content or an event tag representing a vocal event, with precise placement and, for continuous events, appropriate span marking. For example:

\textcolor[HTML]{4635DE}{
\textit{Input audio:} [audio clip](A person whispers ``hello'', then laughs.)}
\textcolor[HTML]{D04699}{
\textit{Output transcription:} \texttt{<whispering>} hello \texttt{</whispering>} \texttt{[laughs]}}

\section{WESR-Bench}
In this section, we present WESR-Bench, detailing the construction of our expert-annotated dataset and the standardized word-level evaluation metrics.

\subsection{Data Construction}
\label{sec:retrieval}

\paragraph{Data Curation} We collect web-scale audio data from diverse sources, including movies, TV dramas, podcasts, and audiobooks. To ensure audio quality, we employ MossFormer2~\cite{zhao2024mossformer2} for denoising and filter out samples with DNSMOS below 2.0, retaining only high-quality recordings suitable for rigorous evaluation.

\paragraph{Hybrid Retrieval Strategy} Constructing an evaluation set rich in vocal events via random sampling is impractical due to the sparsity of such events in general speech. To address this, we develop a hybrid retrieval mechanism to efficiently mine target samples. We leverage two complementary models: 
1) BEATs~\cite{chen2022beats}, a pretrained acoustic encoder that captures non-verbal audio patterns, used for audio-based vector search; 
2) AF-CLAP~\cite{ghosh2025audioflamingo2}, an enhanced CLAP-style encoder with strong audio-text alignment for audio events, used for text-based retrieval over ASR transcripts. 
For each label class, we design three representative queries in the text modality and three in the audio modality. By retrieving utterances with high similarity to these queries, we obtain a candidate set of 1,297 utterances for expert annotation in WESR-Bench.

\begin{figure}[t]
    \centering
    \includegraphics[width=1\linewidth]{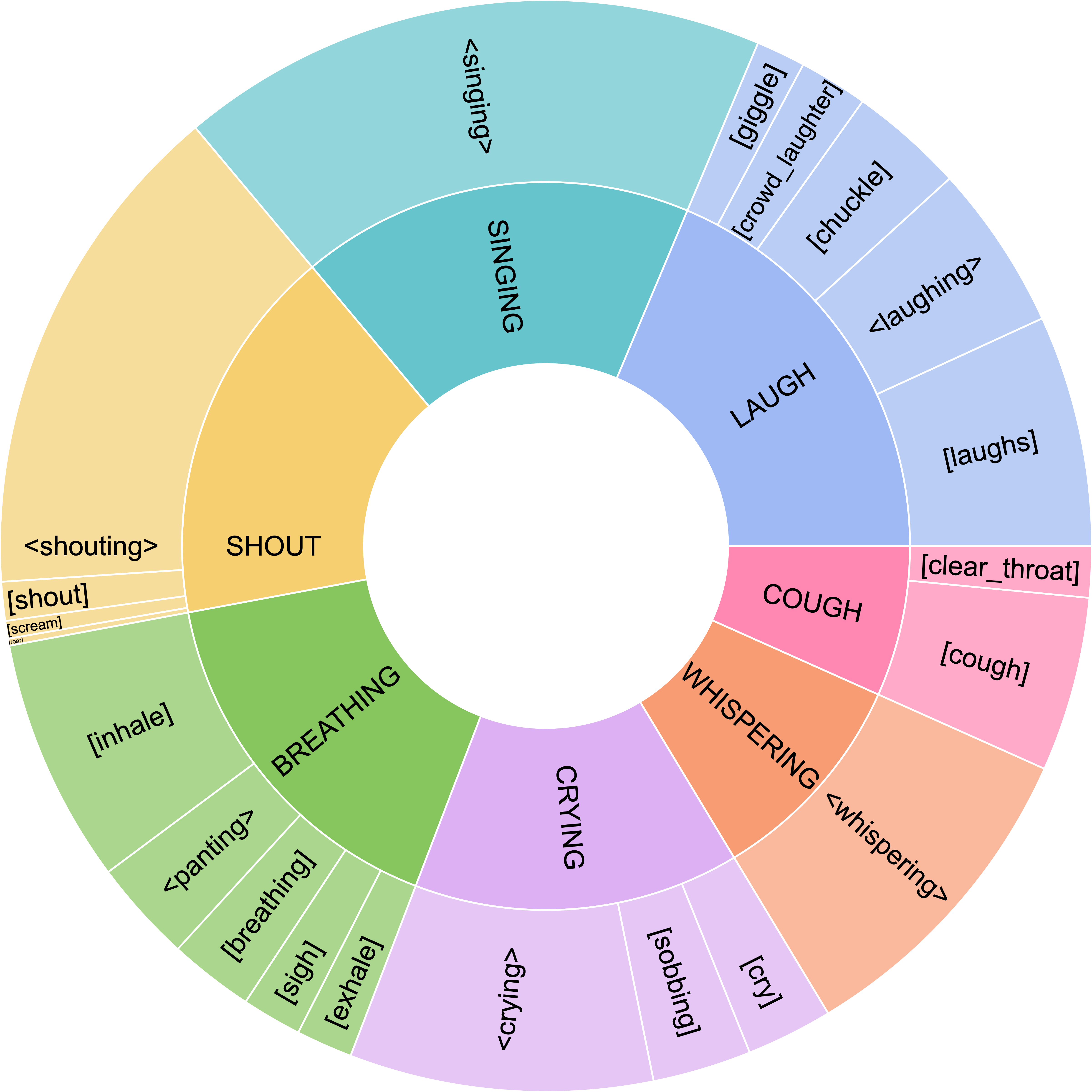}
    \caption{Tag distribution in WESR-Bench. The inner ring shows major event categories, each represented by a distinct color. The outer ring displays the specific tag instances within each category.}
    \label{fig:wesr-bench-sunburst}
\end{figure}

\subsection{Human Annotation}

To ensure the benchmark serves as a reliable gold standard, we implement a rigorous human annotation process. We recruit three annotators who completed a mandatory training program and passed a qualification test before participating. The training includes a tutorial on event ontology and a guided exercise with feedback (annotation guidelines and interface are provided in Appendix~\ref{app:guidelines}). Annotators were compensated at a rate of \$30 per hour of audio annotated. During the annotation phase, they worked independently to insert event tags without referencing retrieval results in Section~\ref{sec:retrieval}. Finally, a senior expert reviewed all samples to verify boundary precision and classification correctness to ensure quality.

\subsection{Dataset Analysis}
\paragraph{Statistics}
After filtering out utterances where annotators found no valid vocal events or ambiguous temporal boundaries, the final benchmark comprises 927 verified utterances. The distribution of tag categories is shown in Figure~\ref{fig:wesr-bench-sunburst}, and per-utterance statistics are detailed in Table~\ref{tab:tag_distribution}. Notably, 29.13\% of utterances contain multiple tags, and 24.60\% feature multiple distinct event types, indicating a meaningful presence of diverse vocal events within single utterances. The corpus contains 1,918 tag occurrences in total, with a split of 58.8\% continuous events and 41.2\% discrete events. The dataset comprises 58\% Chinese and 42\% English by duration.

\begin{table}[b]
\centering
\footnotesize
\renewcommand{\arraystretch}{1.2}
\begin{tabularx}{\linewidth}{>{\raggedright\arraybackslash}X>{\centering\arraybackslash}p{1.2cm}>{\centering\arraybackslash}p{1.2cm}>{\centering\arraybackslash}p{1.2cm}}
\toprule
\textbf{Metric} & \textbf{\# Tags} & \textbf{\# Utt.} & \textbf{\%} \\
\midrule
\multirow{3}{*}{Total Tags} 
& 1 & 657 & 70.87 \\
& 2 & 184 & 19.85 \\
& $\geq 3$ & 86 & 9.28 \\
\midrule
\multirow{3}{*}{Unique Tags} 
& 1 & 699 & 75.40 \\
& 2 & 180 & 19.42 \\
& $\geq 3$ & 48 & 5.18 \\
\bottomrule
\end{tabularx}
\caption{Distribution of utterances by tag statistics. \textbf{Total Tags}: all event tags in an utterance (continuous pairs \texttt{< >...</>} counted as one). \textbf{Unique Tags}: distinct event categories (repeated tags counted once). ``\# Utt.'' indicates the number of utterances.}
\label{tab:tag_distribution}
\end{table}

\paragraph{Case Study}
Table~\ref{tab:casestudy} presents cases in our proposed WESR-Bench, illustrating the diversity of events of our word-level event-speech recognition task. We showcase both discrete events, such as \texttt{[clear\_throat]} and \texttt{[laughs]}, as well as continuous events, such as \texttt{<singing>} and \texttt{<whispering>}. We also demonstrate that mixed event tags can occur in one utterance. 
These examples demonstrate the ability of our method to capture fine-grained word-level boundaries and provide rich annotations for a wide range of vocal event types.

\subsection{Evaluation Protocol}
\label{sec:eval}

\begin{figure}[t]
    \centering
    \includegraphics[width=1\linewidth]{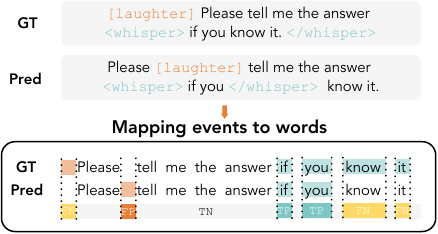}
    \caption{Illustration of Steps 2 and 3 of the WESR evaluation method after Event-Preserving Alignment. Event tags are extracted and mapped to their corresponding word or inter-word positions for metrics calculation.}
    \label{fig:eval}
\end{figure}

To rigorously assess the performance of WESR, we focus on two key aspects: whether all occurring vocal events are successfully detected and whether their predicted positions are accurate. Since direct comparison of event labels can lead to misalignment when predicted text differs from the ground truth, we first align the hypothesis and reference transcripts at the word level. Subsequently, we map and compare the event tags based on this alignment to calculate the final metrics as shown in Figure~\ref{fig:eval}. The detailed process is described below:

\paragraph{Step 1. Event-Preserving Alignment}

To align the hypothesis and reference sequences without altering event tags, an event-preserving alignment procedure is applied. First, event tags are temporarily removed from the reference to obtain plain text for alignment, while being preserved in the hypothesis. Then, using SequenceMatcher, we align the hypothesis to the reference, generating edit operations (insert, delete, replace). Finally, the operations are executed that: insertions add reference words to the hypothesis; deletions remove only non-event words; for replacements, event tags within the replaced segment are extracted and preserved, the segment is replaced with the reference text, and the extracted event tags are re-inserted at the most similar positions in the new segment. The pseudocode for this process is shown in Appendix~\ref{app:alignment}.

\paragraph{Step 2. Mapping Events to Words}

To precisely locate each label, we introduce the concepts of ``word positions'' and ``inter-word positions.'' Continuous events are assigned to all words within their span, indicating that the event persists across those words; discrete events are assigned to the positions between two words, marking a momentary event occurrence. Thus, for a sequence of $N$ words, there are $2N+1$ possible positions: $N$ word positions and $N+1$ inter-word positions (including start and end). This design captures both continuous intervals and exact insertion points, enabling unified evaluation of different label types.

\paragraph{Step 3. Metrics Calculation}
With the aligned sequence and unified word/inter-word positions, we compute true positives~($TP$), false positives~($FP$), and false negatives~($FN$) for each label type. $TP$ are positions correctly labeled with the event, $FP$ are positions incorrectly labeled with an event that should not be there, and $FN$ are positions that should have an event label but are missing or incorrect. Note that true negatives are not counted, as most positions in conversational speech do not contain vocal events.

\section{Building a Strong Baseline for WESR}
In this section, we present a strong baseline for WESR by constructing the \textsc{WESR-Train} dataset and training WESR-specialized models. We then evaluate these models against recent open-source ALMs and commercial APIs on our WESR-Bench.

\begin{table*}[t]
\centering
\scriptsize
\setlength{\tabcolsep}{3pt} 
\renewcommand{\arraystretch}{1.2}
\begin{tabularx}{\textwidth}{l >{\centering\arraybackslash}X >{\centering\arraybackslash}X >{\centering\arraybackslash}X >{\centering\arraybackslash}X| >{\centering\arraybackslash}X >{\centering\arraybackslash}X >{\centering\arraybackslash}X }
\toprule
\multirow{2}{*}{\textbf{Tag}} &
\textbf{Kimi-Audio} &
\textbf{Qwen3-Omni} &
\textbf{Gemini-2.5-Pro} &
\textbf{Gemini-3-Pro} &
\textbf{WESR-Whisper} &
\textbf{WESR-Kimi} &
\textbf{WESR-Qwen} \\
& P / R / F1 (\%) & P / R / F1 (\%) & P / R / F1 (\%) & P / R / F1 (\%) & P / R / F1 (\%) & P / R / F1 (\%) & P / R / F1 (\%) \\
\midrule
\texttt{<crying>} & 28.9 / 04.6 / 08.0 & 55.3 / 11.7 / 19.4 & 59.4 / 80.3 / 68.3 & 66.2 / 73.7 / 69.7 & 65.1 / 89.4 / \textbf{75.3} & 60.7 / 90.2 / \underline{72.5} & 60.8 / 92.1 / 73.3 \\
\texttt{<laughing>} & 00.4 / 00.3 / 00.4 & 03.8 / 12.5 / 05.8 & 22.5 / 17.3 / 19.6 & 30.8 / 25.6 / \textbf{28.0} & 29.8 / 12.5 / 17.6 & 26.3 / 15.7 / 19.7 & 35.4 / 22.4 / \underline{27.5} \\
\texttt{<panting>} & \phantom{0}0\phantom{.0} / \phantom{0}0\phantom{.0} / \phantom{0}0\phantom{.0} & \phantom{0}0\phantom{.0} / \phantom{0}0\phantom{.0} / \phantom{0}0\phantom{.0} & 39.8 / 30.5 / \underline{34.5} & 37.8 / 20.5 / 26.6 & 32.4 / 32.9 / 32.7 & 34.5 / 24.5 / 28.6 & 38.0 / 36.9 / \textbf{37.5} \\
\texttt{<shouting>} & 31.5 / 17.2 / 22.2 & 73.0 / 41.6 / 53.0 & 77.9 / 53.7 / 63.6 & 62.8 / 68.4 / \underline{65.5} & 72.1 / 61.7 / \textbf{66.5} & 65.4 / 55.7 / 60.2 & 72.5 / 59.2 / 65.2 \\
\texttt{<singing>} & 57.1 / 00.5 / 00.9 & 96.3 / 75.9 / 84.9 & 98.5 / 87.2 / 92.5 & 97.3 / 81.6 / 88.7 & 99.6 / 92.4 / \underline{95.9} & 99.4 / 93.7 / \textbf{96.5} & 98.8 / 93.0 / 95.8 \\
\texttt{<whispering>} & \phantom{0}0\phantom{.0} / \phantom{0}0\phantom{.0} / \phantom{0}0\phantom{.0} & 59.5 / 10.8 / 18.3 & 96.4 / 14.1 / 24.7 & 94.0 / 37.1 / 53.2 & 85.6 / 64.4 / 73.5 & 84.2 / 67.7 / \underline{75.1} & 85.9 / 69.9 / \textbf{77.1} \\
\texttt{[breathing]} & 10.3 / 06.2 / \underline{07.8} & 03.3 / 02.1 / 02.6 & 05.0 / 02.1 / 02.9 & 16.4 / 22.9 / \textbf{19.1} & \phantom{0}0\phantom{.0} / \phantom{0}0\phantom{.0} / \phantom{0}0\phantom{.0} & 03.6 / 02.1 / 02.6 & 02.3 / 02.1 / 02.2 \\
\texttt{[chuckle]} & \phantom{0}0\phantom{.0} / \phantom{0}0\phantom{.0} / \phantom{0}0\phantom{.0} & 40.0 / 03.1 / 05.7 & 12.8 / 32.3 / 18.3 & 20.0 / 12.3 / 15.2 & 16.8 / 52.3 / \underline{25.5} & 20.9 / 52.3 / \textbf{29.8} & 16.6 / 53.8 / 25.4 \\
\texttt{[clear\_throat]} & 38.9 / 48.3 / 43.1 & 30.4 / 48.3 / 37.3 & 44.2 / 65.5 / 52.8 & 51.4 / 62.1 / 56.2 & 67.9 / 65.5 / 66.7 & 74.1 / 69.0 / \textbf{71.4} & 65.6 / 72.4 / \underline{68.9} \\
\texttt{[cough]} & 39.7 / 31.3 / 35.0 & 64.6 / 42.4 / 51.2 & 51.4 / 55.6 / 53.4 & 65.4 / 52.0 / 58.0 & 74.4 / 67.7 / \textbf{70.9} & 71.8 / 56.6 / \underline{63.3} & 69.6 / 55.6 / 61.8 \\
\texttt{[crowd\_laughter]} & \phantom{0}0\phantom{.0} / \phantom{0}0\phantom{.0} / \phantom{0}0\phantom{.0} & \phantom{0}0\phantom{.0} / \phantom{0}0\phantom{.0} / \phantom{0}0\phantom{.0} & 66.7 / 15.4 / 25.0 & 29.0 / 23.1 / \underline{25.7} & 42.9 / 15.4 / 22.6 & 44.4 / 10.3 / 16.7 & 56.2 / 23.1 / \textbf{32.7} \\
\texttt{[cry]} & 11.4 / 08.3 / 09.6 & 00.5 / 02.1 / 00.9 & 16.2 / 22.9 / \textbf{19.0} & 05.9 / 02.1 / 03.1 & 100 / 08.3 / 15.4 & 83.3 / 10.2 / \underline{18.2} & 28.6 / 04.2 / 07.3 \\
\texttt{[exhale]} & 03.2 / 03.1 / 03.2 & 10.0 / 28.1 / \underline{14.8} & 06.5 / 06.2 / 06.3 & 14.3 / 09.4 / 11.3 & 20.0 / 09.4 / 12.8 & 31.6 / 18.8 / \textbf{23.5} & 15.0 / 09.4 / 11.5 \\
\texttt{[giggle]} & 16.7 / 10.7 / 13.0 & 02.4 / 07.1 / 03.5 & 10.5 / 28.6 / 15.4 & 20.0 / 22.2 / 21.1 & 25.6 / 39.3 / \underline{31.0} & 28.6 / 35.7 / \textbf{31.7} & 19.6 / 32.1 / 24.3 \\
\texttt{[inhale]} & 10.2 / 07.3 / 08.5 & 11.1 / 01.5 / 02.6 & 08.7 / 09.5 / 09.1 & 06.0 / 33.6 / 10.1 & 09.6 / 06.6 / 07.8 & 13.5 / 09.4 / \underline{11.1} & 14.8 / 09.5 / \textbf{11.6} \\
\texttt{[laughs]} & 20.2 / 58.8 / 30.0 & 33.3 / 38.2 / 35.6 & 33.8 / 35.9 / 34.8 & 30.8 / 61.8 / 41.1 & 45.5 / 42.0 / \underline{43.7} & 49.6 / 45.6 / \textbf{47.5} & 44.5 / 40.5 / 42.4 \\
\texttt{[roar]} & \phantom{0}0\phantom{.0} / \phantom{0}0\phantom{.0} / \phantom{0}0\phantom{.0} & \phantom{0}0\phantom{.0} / \phantom{0}0\phantom{.0} / \phantom{0}0\phantom{.0} & 16.7 / 25.0 / 20.0 & \phantom{0}0\phantom{.0} / \phantom{0}0\phantom{.0} / \phantom{0}0\phantom{.0} & 50.0 / 25.0 / \textbf{33.3} & 50.0 / 25.0 / \textbf{33.3} & 20.0 / 25.0 / \underline{22.2} \\
\texttt{[scream]} & 04.2 / 10.0 / 05.9 & \phantom{0}0\phantom{.0} / \phantom{0}0\phantom{.0} / \phantom{0}0\phantom{.0} & 10.0 / 20.0 / 13.3 & 08.7 / 20.0 / 12.1 & 21.1 / 40.0 / \underline{27.6} & 08.3 / 10.0 / 09.1 & 23.5 / 40.0 / \textbf{29.6} \\
\texttt{[shout]} & 10.0 / 27.3 / 14.6 & 02.4 / 09.1 / 03.7 & 20.0 / 13.6 / 16.2 & 25.0 / 22.7 / \textbf{23.8} & 40.0 / 09.1 / 14.8 & 40.0 / 09.1 / 14.8 & 50.0 / 13.6 / \underline{21.4} \\
\texttt{[sigh]} & 07.4 / 05.9 / 06.6 & 09.9 / 23.5 / 13.9 & 12.1 / 47.1 / 19.3 & 26.8 / 44.1 / 33.3 & 27.0 / 58.8 / \underline{37.8} & 35.9 / 67.6 / \textbf{46.9} & 27.3 / 61.8 / \underline{37.8} \\
\texttt{[sobbing]} & 05.4 / 19.3 / 08.5 & 00.6 / 03.5 / 01.0 & 12.8 / 40.4 / 19.5 & 09.6 / 42.1 / 15.7 & 12.6 / 50.9 / 20.2 & 14.1 / 48.2 / \underline{21.9} & 13.7 / 57.9 / \textbf{22.1} \\
\midrule
\rowcolor{gray!12}
Micro avg. & 19.4 / 04.5 / 07.3 & 47.8 / 27.0 / 34.5 & 64.3 / 46.4 / 53.9 & 63.8 / 54.7 / 58.9 & 72.5 / 68.9 / \underline{70.6} & 71.3 / 69.7 / 70.5 & 71.2 / 71.7 / \textbf{71.4} \\
\rowcolor{gray!12}
Macro avg. & 14.1 / 12.3 / 10.3 & 23.6 / 17.2 / 16.9 & 34.4 / 33.5 / 29.9 & 34.2 / 35.1 / 32.3 & 44.7 / 40.2 / \underline{37.7} & 44.8 / 38.9 / 37.8 & 40.9 / 41.6 / \textbf{38.0} \\
\bottomrule
\end{tabularx}
\caption{Performance of various models on all vocal event categories in WESR-Bench. The first four columns show results using 2-shot prompting (Kimi-Audio, Qwen3-Omni, Gemini-2.5-Pro, Gemini-3-Pro), while the last three columns show results for our WESR-trained models (WESR-Whisper, WESR-Kimi, WESR-Qwen). ``Micro avg.'' is computed by directly averaging across all samples, while ``Macro avg.'' is computed by averaging per-category performance. The best and second-best F1 scores in each row are bolded and underlined, respectively.}
\label{tab:main_result}
\end{table*}

\subsection{WESR-Train}
To facilitate training for word-level event-speech recognition, we construct a large-scale weakly labeled dataset called \textsc{WESR-Train}. 

\paragraph{Data Collection and Annotation}
For web-sourced data, we employ a similar data curation and hybrid retrieval strategy as described in Section~\ref{sec:retrieval}. We use different text/audio queries from those used in constructing WESR-Bench to retrieve diverse samples. To ensure fair evaluation, we perform deduplication against WESR-Bench to prevent any overlap. The collected data is then automatically annotated using Gemini~\footnote{\texttt{gemini-2.5-pro}}, the latest version available at the time of data collection, to generate word-level event annotations (see Appendix~\ref{app:prompt} for the prompt details). 

\paragraph{Adaptation from other datasets}
For open-source datasets, we incorporate word-level vocal event data from NonverbalTTS, NVSpeech-170k, NonVerbalSpeech-38K, and SMIIP-NV, adapting their annotations to our format. We normalize their vocal event tags to match our WESR taxonomy through careful mapping, removing audio with tags outside our taxonomy. To ensure mapping quality, we conduct a manual review of the mapped annotations, with particular attention to NonVerbalSpeech-38K, which uses different continuous event definitions than ours. We also observe that audios annotated with continuous \texttt{<B>...</B>} tags in NonVerbalSpeech-38K contain significant annotation errors and thus exclude them from our training set. The final data distribution of WESR-Train is shown in Table~\ref{tab:dataset_stats}.

\begin{table}[h]
\centering
\normalsize
\setlength{\tabcolsep}{4.2pt} 
\footnotesize
\renewcommand{\arraystretch}{1.2}
\begin{tabularx}{\linewidth}{X r r}
\toprule
\textbf{Data Source} & \textbf{Dur. (h)} & \textbf{Language} \\
\midrule
NonverbalTTS & 14 & EN \\
NVSpeech-170k & 332 & EN, ZH \\
NonVerbalSpeech-38K & 87 & EN, ZH \\
SMIIP-NV & 35 & ZH \\
Gemini-annotated & 1,299 & EN, ZH \\
\midrule
\rowcolor{gray!20}
\textsc{WESR-Train} & \textbf{1,767} & EN, ZH \\
\bottomrule
\end{tabularx}
\caption{Composition of WESR-Train dataset. ``Dur.'' denotes the total duration.}
\label{tab:dataset_stats}
\end{table}

\subsection{Adaptation across Backbones}

Leveraging the rich annotations provided by WESR-Train, we apply supervised fine-tuning to three distinct backbones: Whisper-Large-v3 (1.5B)~\citep{radford2023whisper}, Kimi-Audio-7B-Instruct (7B)~\citep{kimiteam2025kimiaudiotechnicalreport}, and Qwen3-Omni-Instruct (30B)~\citep{Qwen3-Omni}, and evaluate their performance on WESR-Bench. Our experiment setup is detailed in Appendix~\ref{app:training}.

As shown in Table~\ref{tab:main_result}, our approach achieves consistent performance across different architectures, with Macro F1 scores ranging from 37.7\% to 38.0\% despite significant variations in model size (1.5B to 30B parameters). This demonstrates that our training approach generalizes well across diverse model architectures and scales.

For the subsequent evaluation and comparison, we select our fine-tuned Qwen3-Omni as the representative model, given its consistently strong performance across different event categories.

\subsection{Comparison with Other Baselines}

To comprehensively evaluate our proposed methods, we compared their performance with other baselines of several representative audio language models (ALMs) on WESR-Bench. ALMs considered in our evaluation are as follows: 1) Newest open-source ALMs such as Kimi-Audio~\cite{kimiteam2025kimiaudiotechnicalreport}, MiMo-Audio~\cite{zhang2025mimo}, and Qwen3-Omni~\cite{Qwen3-Omni}. 2) Commercial APIs, such as Gemini~\footnote{\texttt{gemini-2.5-pro, gemini-3-pro}} and GPT~\footnote{\texttt{gpt-4o-audio-preview}}. We prompt these models using the same instruction (see Appendix~\ref{app:prompt}), which includes two in-context examples illustrating the use of discrete and continuous tags. The detailed results are presented in Table~\ref{tab:main_result}.

As shown in Table~\ref{tab:main_result}, different models exhibit substantial differences in non-verbal event detection. Overall, our Qwen3-Omni fine-tuned WESR model achieves the highest F1 scores across all categories, reaching a macro F1 of 38.0\%, with absolute improvements of 21.1\% over Qwen3-Omni and 8.1\% over Gemini-2.5-Pro. This improvement is particularly pronounced in challenging categories such as \texttt{<panting>}, \texttt{<whispering>}, and various laughter-related tags (\texttt{[chuckle]}, \texttt{[giggle]}, \texttt{[laughs]}), where few-shot models often achieve near-zero recall but fine-tuned models attain substantially higher detection rates. For instance, WESR-whisper achieves 64.4\% recall on \texttt{<whispering>} compared to Gemini-3-Pro's 37.1\%, and 52.3\% on \texttt{[chuckle]} versus 12.3\%. These results suggest that while large multimodal models possess some inherent capability for vocal event recognition through prompting, systematic exposure to labeled training data remains critical for achieving robust and reliable performance across diverse paralinguistic phenomena. 

During evaluation, we also notice that GPT-4o and MiMo-Audio frequently refused to respond to the WESR prompt. In the rare cases where it did produce an output, it failed to insert any event tags into the transcription, rendering them unsuitable for our WESR task.

We also report aggregated results in Supplementary Table \ref{tab:aggregated_result}, where fine-grained tags are grouped into broader vocal event categories (e.g., grouping \texttt{[clear\_throat]} and \texttt{[cough]} into \textit{Cough}, detailed in Supplementary Table~\ref{tab:label_aggregation}). The aggregated metrics further confirm the ability of our model on distinguishing different vocal event categories. We also observe highly synchronized performance trends: all models achieve their best performance on distinct, sustained events such as \textit{Singing} (F1 > 94\%) and \textit{Whispering} (F1 >75\%), while consistently facing challenges with subtle, low-energy acoustic features like \textit{Breathing}.

\subsection{Comparison on Discrete vs. Continuous Events}

Continuous events (e.g., \texttt{<crying>...</crying>}) present a greater challenge than discrete events (e.g., \texttt{[cough]}). The model must not only infer the correct event type but also determine accurate span boundaries. To study all baselines' ability on discrete and continuous events, we calculate F1 scores separately for discrete and continuous events. Note that continuous F1 scores are generally higher than discrete ones: continuous tags allow partial credit through token-level overlap when boundaries are slightly misaligned (see Section~\ref{sec:eval}), whereas discrete tags require exact position matching.

\begin{figure}[t]
    \centering
    \includegraphics[width=\linewidth]{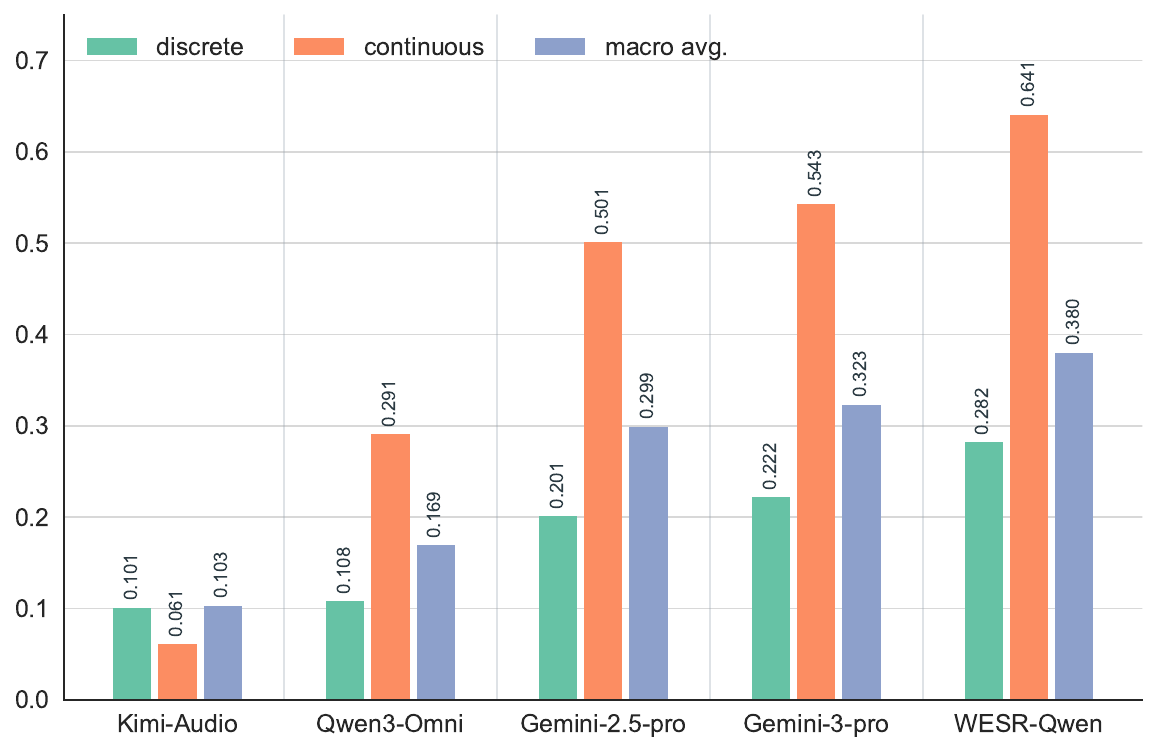}
    \caption{Performance comparison between continuous and discrete event tags on WESR-Bench.}
    \label{fig:continuous_tag_bars}
\end{figure}

As shown in Figure~\ref{fig:continuous_tag_bars}, open-source models struggle significantly: Kimi-Audio achieves only 0.101 / 0.061 F1 (discrete / continuous), while Qwen3-Omni reaches 0.108 / 0.291. Proprietary models perform better, with Gemini-2.5-Pro at 0.201 / 0.501 and Gemini-3.0-Pro at 0.222 / 0.543. WESR-Qwen substantially outperforms all baselines on both event types, achieving 0.282 F1 on discrete events (+27\% over Gemini-3.0-Pro) and 0.641 F1 on continuous events (+18\% improvement). This demonstrates that our specialized training effectively enhances both precise event localization and accurate span boundary prediction.

\subsection{Impact on ASR Performance}
To explore the impact of WESR to the underlying ALM's ASR ability, we conduct experiments on Common Voice 15~\citep{ardila-etal-2020-commonvoice} test set, a widely used large-scale ASR benchmark. We compare the word error rate (WER) of the original model and our WESR-fine-tuned models on both English (en) and Chinese (zh-CN) test sets. Since Common Voice does not include vocal event annotations, we exclude vocal event tags from the WER computation to ensure a fair comparison. The results are shown in Table~\ref{tab:wer}.

\begin{table}[h!]
\footnotesize
    \centering
    \begin{tabular}{ccc}
        \toprule
        \textbf{Model} & \textbf{en} & \textbf{zh-CN} \\
        \midrule
        Qwen3-Omni & 7.2 & 6.0\\
        WESR-Qwen & 8.6 & 7.2\\
        \bottomrule
    \end{tabular}
    \caption{Word error rate (WER, \%) comparison of the original Qwen3-Omni and our WESR fine-tuned model on CommonVoice 15 test set (en and zh-CN). Lower WER indicates better performance.}
    \label{tab:wer}
\end{table}

Results show that our WESR fine-tuned model maintains competitive ASR performance with only modest increases in WER, demonstrating that WESR can be integrated without significantly compromising transcription accuracy.

\section{Conclusion}

In this work, we introduce the Word-level Event-Speech Recognition (WESR) task. We formalize a rigorous taxonomy categorizing 21 vocal events into \textit{discrete} and \textit{continuous} types, establishing a comprehensive framework for this task. We develop \textsc{WESR-Bench}, an expert-annotated benchmark with a novel position-aware protocol that disentangles ASR errors from event detection, enabling precise localization measurement. We establish a strong baseline for WESR by constructing WESR-Train and training specialized models that, across different parameter scales, outperform both open-source audio-language models and commercial APIs while maintaining ASR quality. We believe WESR will serve as a foundational resource for future research in modeling rich, real-world auditory scenes.

\section*{Limitations}
We acknowledge that our work may have the following limitations: 1) Language Coverage: Our system is currently verified only for English and Chinese. While the underlying methodology is language-agnostic, its generalization capabilities to languages with different morphological structures or lower resource availability remain unverified. 2) Resource Intensity: The construction of our high-quality dataset relies on the commercial Gemini API and expert annotation. While this ensures data quality, it presents a trade-off in terms of cost-efficiency and scalability. This may pose challenges for reproducibility in resource-constrained environments. 

\section*{Ethical Considerations}
We prioritize ethical standards throughout our data construction and modeling processes. Our datasets utilize publicly available audio under fair use principles. We ensured the welfare of our annotators through fair compensation (\$30/h). 
Due to our large-scale annotation process, comprehensive manual verification of all data instances is not feasible. As such, the dataset may inadvertently include inappropriate content. We emphasize that any content appearing in the source audio or annotations does NOT reflect the perspectives, beliefs, or endorsements of the authors. We release our code, data, and models solely for academic use.

% \section*{Acknowledgments}

\bibliography{custom}

\clearpage
\appendix
\label{sec:appendix}

\section{Annotation Details}
\label{app:guidelines}
\begin{figure*}[h]
\centering
\includegraphics[width=1\linewidth]{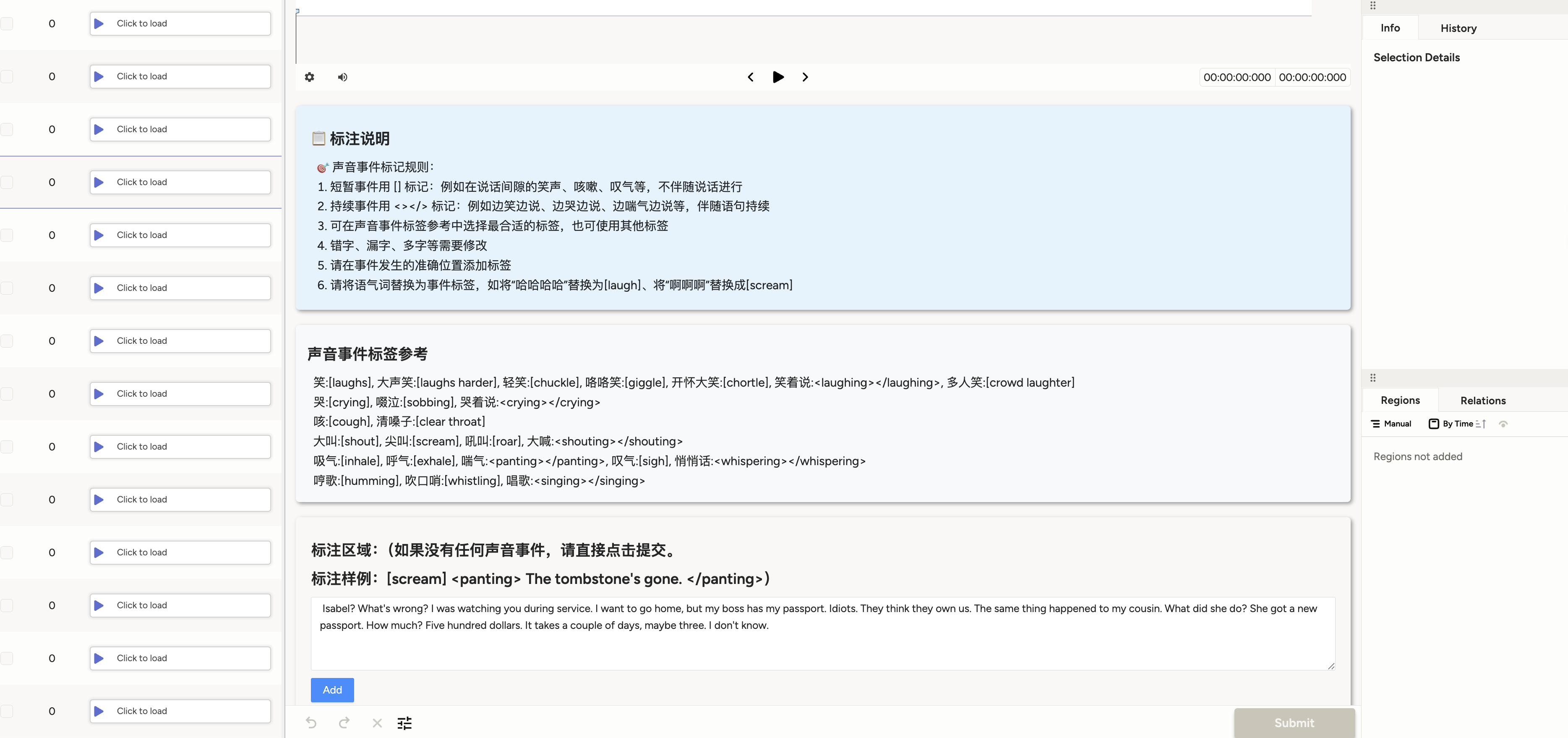}
\caption{The labeling interface for human annotation.}
\label{fig:labeling_screenshot}
\end{figure*}

Table~\ref{tab:guideline} shows instructions on vocal event taxonomy and examples shown to annotators. Figure~\ref{fig:labeling_screenshot} shows the annotation page.

\section{Evaluation Details}
\label{app:alignment}
\begin{algorithm}[h]
\caption{Event-Preserving Alignment}
\label{alg:event_alignment}
\begin{algorithmic}[1]
\REQUIRE Hypothesis sequence $H$, Reference sequence $R$
\ENSURE Aligned hypothesis sequence $H'$

\STATE $R_{text} \gets$ Remove all event tags from $R$
\STATE $H_{split} \gets$ Split $H$ into list (preserving event tags as separate elements)
\STATE $R_{split} \gets$ Split $R_{text}$ into list

\STATE $ops \gets$ SequenceMatcher($H_{split}$, $R_{split}$) \COMMENT{Get edit operations}

\STATE $H' \gets H_{split}$

\FOR{each operation $op$ in $ops$}
    \IF{$op$ is INSERT}
        \STATE Insert corresponding words from $R_{split}$ into $H'$
    \ELSIF{$op$ is DELETE}
        \STATE Delete only non-event words from $H'$
    \ELSIF{$op$ is REPLACE}
        \STATE $events \gets$ Extract event tags from replaced segment in $H'$
        \STATE Replace segment in $H'$ with corresponding text from $R_{split}$
        \STATE Re-insert $events$ at most similar positions in new segment
    \ENDIF
\ENDFOR

\RETURN $H'$ \COMMENT{Aligned sequence matching reference in non-event content}
\end{algorithmic}
\end{algorithm}

\section{Training Details}
\label{app:training}
For Whisper, we fine-tuned whisper-large-v3 using a learning rate of $1 \times 10^{-5}$, global batch size of 8, warmup steps ratio of 0.1, and trained for 3 epochs on H100*8 for 4 hours.

For Kimi, fine-tuned Kimi-Audio-7B-Instruct using a learning rate of $1 \times 10^{-6}$, packing sequence length of 4,096 tokens, and trained for 3 epochs on H100*8 for 5 hours.

For Qwen3-Omni, we fine-tuned Qwen3-Omni-30B-A3B-Instruct with a learning rate of $1 \times 10^{-6}$, packing sequence length of 4,096 tokens, and trained for 3 epochs on H200*8 for 5 hours.

All hyperparameters were determined by conducting preliminary experiments with three different learning rates on a validation set, which was randomly sampled from the training set. Vocal event tags are added as special tokens.

\section{Prompt}
\label{app:prompt}

\begin{lstlisting}[language=,frame=single]
"""
You are an expert in recognizing special patterns in audio. Please perform the following task:

Transcribe text from the given audio, and insert the following fine-grained vocal event tags into the precise position to the text according to what is detected in the audio.
The supported tags are: [laughs], [chuckle], [giggle], <laughing></laughing>, [crowd_laughter], [crying], [sobbing], <crying></crying>, [cough], [clear_throat], [shout], [scream], [roar], <shouting></shouting>, <whispering></whispering>, [inhale], [exhale], <panting></panting>, [breathing], [sigh], <singing></singing>.

Tag usage guidelines:

Square brackets [ ]: Insert the tag at the exact point where the event occurs, typically between words or within a sentence. For example:
[inhale] I don't think they saw us.
Angle brackets < >...</>: Use these tags to wrap around specific words or phrases that are spoken while the vocal event is happening. For example:
<laughing> Just like that! </laughing>

Annotation guidelines:

Only add event tags at the exact positions where vocal events occur.
The annotation should be precise to the position where the event occurs.
If there are no events to annotate, the output should be the ASR text only.
There can be audio without any vocal event. Do not insert event tags when there are no vocal events in the audio.
Your final output must only contain the text with event annotations. Do not include any other explanations or formatting.

Example:
"I don't think they saw us. [inhale] Let's keep moving."
or
"<laughing> Just like that! </laughing>"
"""
\end{lstlisting}

\clearpage
\onecolumn
\section{Supplementary Tables}

\begin{table*}[h]
\centering
\small
\newcommand{\tagi}[1]{\texttt{\textcolor{blue}{#1}}}
\begin{tabularx}{\textwidth}{l p{4cm} X r}
\toprule
Tag & Explanation & Annotation Example & Audio Example \\
\midrule
\tagi{[laughs]} & Laughter sound & Then \tagi{[laughs]} that was amazing. & (omitted) \\
\tagi{[chuckle]} & Laugh quietly & I know, right? \tagi{[chuckle]} That's exactly what I thought. & (omitted) \\
\tagi{[giggle]} & Laugh in a light, high-pitched way from amusement & She told me the story and I just \tagi{[giggle]} couldn't stop laughing. & (omitted) \\
\tagi{<laughing>} & Speaking while laughing & Later we didn't even need a model; \tagi{<laughing>}this is it. & (omitted) \\
\tagi{[crowd\_laughter]} & Laughter from multiple people & He always hits me… \tagi{[crowd\_laughter]} Oh my… \tagi{[crowd\_laughter]} … parent-teacher meetings were the hardest… \tagi{[crowd\_laughter]}. & (omitted) \\
\midrule
\tagi{[cry]} & Audible crying sound & Look at this dress… \tagi{[cry]} It looks awful; I won't wear it. & (omitted) \\
\tagi{[sobbing]} & Intermittent sobbing sounds & I can't believe it happened… \tagi{[sobbing]} I just can't. & (omitted) \\
\tagi{<crying>} & Speaking while crying / tearful voice & \tagi{<crying>}I picked it up near Liding Street; that was on their wedding day. & (omitted) \\
\midrule
\tagi{[cough]} & Coughing sound & Thanks \tagi{[cough]} for the two days off. & (omitted) \\
\tagi{[clear\_throat]} & Throat clearing sound & \tagi{[clear\_throat]} The talk was inspiring; let's all share our thoughts. & (omitted) \\
\midrule
\tagi{[scream]} & High-pitched loud vocalization & Help me! \tagi{[scream]} Someone please help! & (omitted) \\
\tagi{[roar]} & Loud, rumbling vocalization & The contest begins! \tagi{[roar]} Let's go! & (omitted) \\
\tagi{[shout]} & Loud cry or call & \tagi{[shout]}Who is it? Open the door! & (omitted) \\
\tagi{<shouting>} & Speaking while shouting & \tagi{<shouting>}Trust me! Please! You have to believe me! & (omitted) \\
\midrule
\tagi{[breathing]} & Respiration or panting sound & \tagi{[breathing]} I owe you so much… only you had nothing… & (omitted) \\
\tagi{[inhale]} & Audible breath intake & I owe you so much \tagi{[inhale]} … only you \tagi{[inhale]} had nothing… & (omitted) \\
\tagi{[exhale]} & Audible breath release & I owe you so much… \tagi{[exhale]} only you… \tagi{[exhale]} had nothing… & (omitted) \\
\tagi{<panting>} & Heavy breathing while speaking & \tagi{<panting>}I can't run anymore… I really can't keep going. & (omitted) \\
\tagi{[sigh]} & Sighing sound & He resented the hereditary illness… \tagi{[sigh]} it was so unfair. & (omitted) \\
\midrule
\tagi{<whispering>} & Speaking in whisper & \tagi{<whispering>}My English is so bad—even simple phrases aren't fluent. & (omitted) \\
\midrule
\tagi{<singing>} & Singing voice & \tagi{<singing>}You wrote me into the script and said you must be the lead. & (omitted) \\
\bottomrule
\end{tabularx}
\vspace{3pt}
\captionof{table}{Annotation guidelines for nonverbal vocal events. Use angle brackets <...> for interval events and square brackets [...] for point events.}
\label{tab:guideline}
\end{table*}

\begin{table*}[h]
\centering
\scriptsize
\setlength{\tabcolsep}{3pt} 
\renewcommand{\arraystretch}{1.2}
\begin{tabularx}{\textwidth}{l >{\centering\arraybackslash}X >{\centering\arraybackslash}X >{\centering\arraybackslash}X >{\centering\arraybackslash}X| >{\centering\arraybackslash}X >{\centering\arraybackslash}X >{\centering\arraybackslash}X }
\toprule
\multirow{2}{*}{\textbf{Aggr. Tag}} &
\textbf{Kimi-Audio} &
\textbf{Qwen3-Omni} &
\textbf{Gemini-2.5-Pro} &
\textbf{Gemini-3-Pro} &
\textbf{WESR-Whisper} &
\textbf{WESR-Kimi} &
\textbf{WESR-Qwen} \\
& P / R / F1 (\%) & P / R / F1 (\%) & P / R / F1 (\%) & P / R / F1 (\%) & P / R / F1 (\%) & P / R / F1 (\%) & P / R / F1 (\%) \\
\midrule
Breathing & 22.6 / 07.8 / 11.6 & 25.7 / 13.7 / 17.9 & 30.8 / 30.9 / 30.9 & 16.6 / 35.2 / 22.5 & 33.6 / 30.7 / 32.1 & 37.9 / 27.8 / 32.0 & 37.0 / 33.8 / 35.3 \\
Cough & 51.8 / 44.5 / 47.9 & 60.7 / 50.8 / 55.3 & 56.2 / 64.1 / 59.9 & 70.5 / 62.2 / 66.1 & 77.2 / 68.8 / 72.7 & 79.4 / 63.3 / 70.4 & 71.8 / 61.7 / 66.4 \\
Crying & 18.9 / 04.6 / 07.4 & 68.9 / 33.9 / 45.5 & 56.4 / 78.8 / 65.8 & 62.0 / 74.4 / 67.7 & 61.7 / 87.8 / 72.5 & 59.2 / 90.1 / 71.5 & 57.7 / 90.4 / 70.4 \\
Laugh & 24.1 / 27.8 / 25.8 & 11.1 / 24.1 / 15.2 & 39.1 / 40.3 / 39.7 & 42.5 / 45.3 / 43.8 & 50.4 / 43.3 / 46.6 & 48.9 / 42.8 / 45.6 & 48.7 / 47.8 / 48.2 \\
Shout & 30.2 / 18.0 / 22.6 & 64.3 / 41.2 / 50.2 & 77.1 / 52.0 / 62.1 & 61.0 / 66.3 / 63.5 & 72.3 / 61.0 / 66.2 & 65.0 / 54.4 / 59.2 & 72.2 / 58.8 / 64.8 \\
Singing & 57.1 / 00.5 / 00.9 & 96.4 / 75.6 / 84.8 & 98.6 / 87.2 / 92.6 & 97.3 / 81.8 / 88.8 & 99.8 / 92.6 / 96.1 & 99.4 / 93.7 / 96.4 & 98.8 / 92.8 / 95.8 \\
Whispering & \phantom{0}0\phantom{.0} / \phantom{0}0\phantom{.0} / \phantom{0}0\phantom{.0} & 59.5 / 10.8 / 18.3 & 96.6 / 14.2 / 24.7 & 94.1 / 37.1 / 53.2 & 85.6 / 64.5 / 73.6 & 84.2 / 67.7 / 75.1 & 85.9 / 70.0 / 77.2 \\
\midrule
\rowcolor{gray!12}
Micro avg. & 26.2 / 05.8 / 09.4 & 58.0 / 32.7 / 41.8 & 68.2 / 48.3 / 56.6 & 66.5 / 57.0 / 61.4 & 75.0 / 71.0 / 73.0 & 73.6 / 71.9 / 72.8 & 73.5 / 73.7 / 73.6 \\
\rowcolor{gray!12}
Macro avg. & 29.3 / 14.7 / 16.6 & 55.2 / 35.7 / 41.0 & 65.0 / 52.5 / 53.7 & 63.4 / 57.5 / 58.0 & 68.6 / 64.1 / 65.7 & 67.7 / 62.8 / 64.3 & 67.5 / 65.0 / 65.4 \\
\bottomrule
\end{tabularx}
\captionof{table}{Performance of various models on aggregated vocal event categories in WESR-Bench. The first four columns show results using few-shot prompting (Kimi-Audio, Qwen3-Omni, Gemini-2.5-Pro, Gemini-3-Pro), while the last three columns show results for models fine-tuned on our WESR-Train corpus (WESR-Whisper, WESR-Kimi, WESR-Qwen).}
\label{tab:aggregated_result}
\end{table*}

\begin{table*}[h]
\centering
\fontsize{8}{9}\selectfont
\begin{tabularx}{\textwidth}{
    >{\raggedright\arraybackslash}l
    >{\raggedright\arraybackslash}X
    >{\raggedright\arraybackslash}c
}
\toprule
Dataset & Tags & Count \\
\midrule
NonverbalTTS & 
\texttt{[breath]}, \texttt{[grunt]}, \texttt{[sniff]}, \texttt{[throat clearing]}, \texttt{[groan]}, \texttt{[sigh]}, \texttt{[snore]}, \texttt{[cough]}, \texttt{[laugh]}, \texttt{[sneeze]}
& 10 \\
\midrule
NVSpeech-170k & 
\texttt{[Breathing]}, \texttt{[Laughter]}, \texttt{[Confirmation-en]}, \texttt{[Uhm]}, \texttt{[Sigh]}, \texttt{[Surprise-ah]}, \texttt{[Surprise-oh]}, \texttt{[Dissatisfaction-hnn]}, \texttt{[Surprise-wa]}, \texttt{[Question-yi]}, \texttt{[Question-ei]}, \texttt{[Cough]}, \texttt{[Question-ah]}, \texttt{[Question-oh]}, \texttt{[Surprise-yo]}, \texttt{[Question-en]}, \texttt{[Shh]}, \texttt{[Crying]}
& 18 \\
\midrule
NonVerbalSpeech-38K & 
\texttt{[snore]}, \texttt{[throatclearing]}, \texttt{[crying]}, \texttt{[breath]}, \texttt{[sniff]}, \texttt{[laughing]}, \texttt{[coughing]}, \texttt{[gasp]}, \texttt{[yawn]}, \texttt{[sigh]}
& 10 \\
\midrule
SMIIP-NV & 
\texttt{[Laughter]}, \texttt{[crying]}, \texttt{[cough]}
& 3 \\
\midrule
Synparaspeech & 
\texttt{[Sigh]}, \texttt{[throat clearing]}, \texttt{[laugh]}, \texttt{[pause]}, \texttt{[tsk]}, \texttt{[gasp]}
& 6 \\
\midrule
MNV-17 & 
\texttt{[Sneezing]}, \texttt{[Clapping]}, \texttt{[Hissing]}, \texttt{[Whistling]}, \texttt{[Clearing Throat]}, \texttt{[Coughing]}, \texttt{[Lip Smacking]}, \texttt{[Exhaling]}, \texttt{[Moaning]}, \texttt{[Panting]}, \texttt{[Sniffling]}, \texttt{[Humming]}, \texttt{[Laughing]}, \texttt{[Applauding]}, \texttt{[Inhaling]}, \texttt{[Chuckling]}, \texttt{[Sighing]}
& 17 \\
\bottomrule
\end{tabularx}
\caption{Summary of non-verbal tags used in prior work.}
\label{tab:prior_tags}
\end{table*}

\begin{table*}[h]
\centering
\fontsize{8}{9}\selectfont
\begin{tabularx}{\textwidth}{l X}
\toprule
Aggregated Category & WESR Event Tags \\
\midrule
\texttt{LAUGH} & 
\texttt{[laughs]}, \texttt{<laughing>}, \texttt{[chuckle]}, \texttt{[giggle]}, \texttt{[crowd laughter]}
\\
\midrule
\texttt{SHOUT} & 
\texttt{[scream]}, \texttt{[roar]}, \texttt{[shout]}, \texttt{<shouting>}
\\
\midrule
\texttt{WHISPERING} & 
\texttt{<whispering>}
\\
\midrule
\texttt{SINGING} & 
\texttt{<singing>}
\\
\midrule
\texttt{BREATHING} & 
\texttt{[inhale]}, \texttt{[exhale]}, \texttt{<panting>}, \texttt{[sigh]}, \texttt{[breathing]}
\\
\midrule
\texttt{COUGH} & 
\texttt{[cough]}, \texttt{[clear\_throat]}
\\
\midrule
\texttt{CRYING} & 
\texttt{<crying>}, \texttt{[sobbing]}, \texttt{[cry]}
\\
\bottomrule
\end{tabularx}
\caption{Label aggregation mapping for non-verbal event tags.}
\label{tab:label_aggregation}
\end{table*}

\clearpage
\section{Supplementary Figures}

\begin{figure*}[h]
    \centering
    \includegraphics[width=0.5\linewidth]{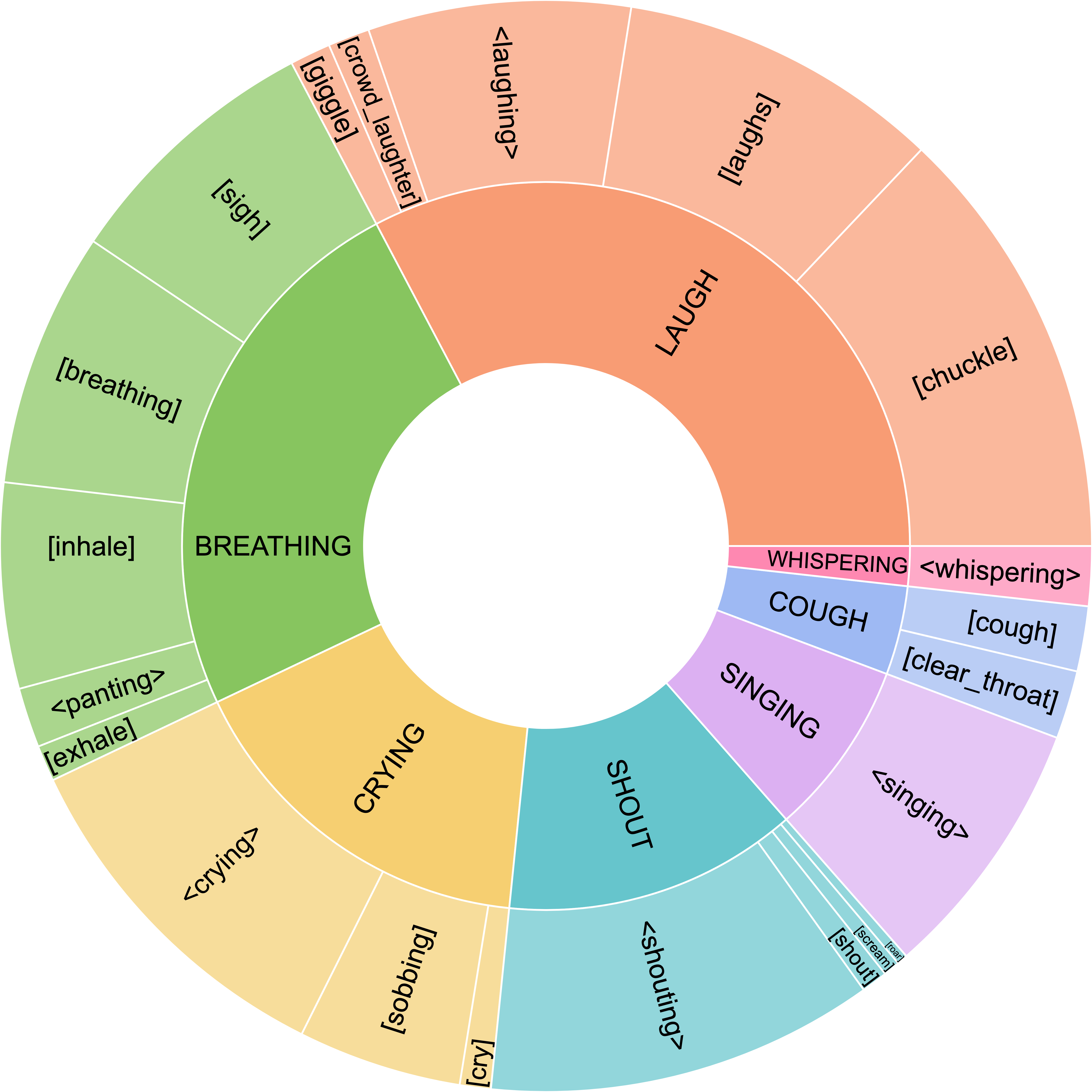}
    \caption{Tag distribution of WESR-Train.}
    \label{fig:wesr-large}
\end{figure*}

\end{document}